\documentclass[letterpaper, 10 pt, conference]{ieeeconf}  

\IEEEoverridecommandlockouts                   
                \usepackage{booktabs}           
\usepackage{graphicx} 
\usepackage{amsmath}
\usepackage{amssymb}  
\usepackage{amsfonts}
\usepackage{algorithmic}
\usepackage{array}
\usepackage{enumerate}
\usepackage{textcomp}
\usepackage{stfloats}
\usepackage{url}
\usepackage{verbatim}
\usepackage{graphicx}
\usepackage{caption}
\usepackage{subcaption}
\usepackage[ruled, vlined, linesnumbered]{algorithm2e}
\usepackage{cite}
\usepackage{amssymb,amsfonts,bm}
\usepackage{amsmath,tikz}
\usepackage{color}
\usepackage{mathtools}
\usepackage{hyperref} 
\hypersetup{
    colorlinks=true,
    linkcolor=black,
    filecolor=magenta,   
    citecolor=black, 
    urlcolor=blue,
    }
\usepackage{rotating}
\usepackage{blkarray}
\usepackage{physics}
\usetikzlibrary{arrows}

\usepackage{amsthm}
\usepackage{comment}
\usepackage{soul}
\usepackage{balance}

\title{\LARGE \bf
A Cost-Effective Cooperative Exploration and Inspection Strategy for Heterogeneous Aerial System
}

\author{Xinhang~Xu,
        Muqing~Cao,
        Shenghai~Yuan,
        Thien Hoang~Nguyen,\\
        Thien-Minh~Nguyen$^*$, \IEEEmembership{Member,~IEEE},
        Lihua~Xie, \IEEEmembership{Fellow,~IEEE}
\thanks{The authors are with School of Electrical and Electronic Engineering, Nanyang Technological University, Singapore 639798, 50 Nanyang Avenue. Email: {\tt xu0021ng@e.ntu.edu.sg, mqcao@ntu.edu.sg, shyuan@ntu.edu.sg,  thienminh.nguyen@ntu.edu.sg,  elxie@ntu.edu.sg}.}
\thanks{This research is supported by the National Research Foundation, Singapore, under its Medium-Sized Center for Advanced Robotics Technology Innovation.}
}

\begin{document}

\maketitle


\begin{abstract}
In this paper, we propose a cost-effective strategy for heterogeneous UAV swarm systems for cooperative aerial inspection. 
Unlike previous swarm inspection works, the proposed method does not rely on precise prior knowledge of the environment and can complete full 3D surface coverage of objects in any shape.
In this work, agents are partitioned into teams, with each drone assign a different task, including mapping, exploration, and inspection. Task allocation is facilitated by assigning optimal inspection volumes to each team, following best-first rules. 
A voxel map-based representation of the environment is used for pathfinding, and a rule-based path-planning method is the core of this approach. 
We achieved the best performance in all challenging experiments with the proposed approach, surpassing all benchmark methods for similar tasks across multiple evaluation trials.

The proposed method is open source at  \url{https://github.com/ntu-aris/caric_baseline} and used as the baseline of the Cooperative Aerial Robots Inspection Challenge at the 62nd IEEE Conference on Decision and Control 2023\cite{CARIC}.

\end{abstract}

\section{Introduction}

Aerial inspection \cite{CARIC, lyu2023vision, cao2023neptune, savva2021icarus} has become the mainstream method for maintaining building integrity, retaining housing value, and improving the safety of owners and inspectors. However, existing methods often rely on a single drone \cite{cao2020online, nguyen2021viral, cao2022direct} flying for an extended period to achieve comprehensive coverage, resulting in prolonged noise interruption to tenants and increased manpower costs to service providers. Various multi-drone swarm inspection solutions \cite{cao2021distributed} have been proposed to address this issue. However, outfitting all drones with expensive LiDAR mapping suite \cite{nguyen2021miliom, nguyen2021liro, nguyen2021viral} for autonomous flight can be costly, and low-cost visual-inertial-based autonomous drones \cite{esfahani2019towards, esfahani2019aboldeepio, wang2017heterogeneous, esfahani2021learning, lyu2022structure, ji2022vision, wang2019fast, wang2018feasible, liu2023non, liu2023multiple, xu2023airvo} are susceptible to perception failures. Exploring heterogeneous drone autonomous inspection solutions is essential to increase efficiency and robustness while maintaining reasonable costs. 
Based on this motivation, the Cooperative Aerial Robots Inspection Challenge (CARIC) was proposed at the 62nd IEEE Conference on Decision and Control in 2023 \cite{CARIC}, which features a system of high-cost mapping and exploring drones, as well as low-cost inspection drones equipped only with cameras, to inspect simulated industrial sites.

\begin{figure}
    \centering
    \includegraphics[width=1\linewidth]{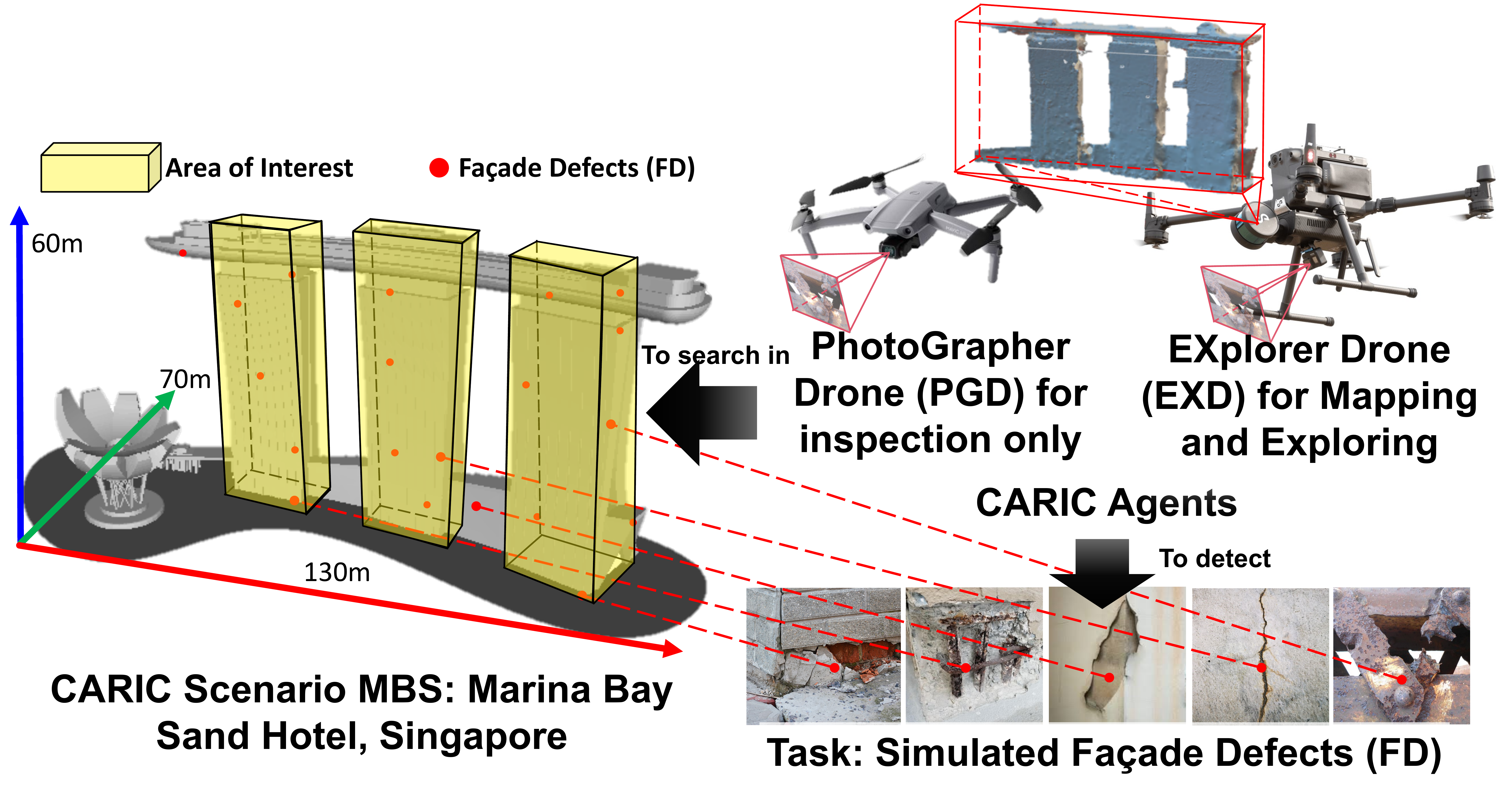}
    \caption{The CARIC conceptual drawing.}
    \label{fig: mission_describe}    
\end{figure}

Heterogeneous drone autonomous inspection without detailed prior information of the environment is a complex problem, consisting of multiple sub-processes such as task assignment, autonomous exploration, and coverage path planning. However, each aspect is a significant challenge on its own \cite{bircher2018receding}.

Firstly, the absence of task assignments in heterogeneous UAV inspection signifies a major domain gap. 
In autonomous exploration, achieving collision-free paths in unknown environments poses a challenge, often requiring online replanning.

The frontier-based method \cite{yamauchi1997frontier} and Next Best View (NBV)-based method\cite{connolly1985determination} are the most intuitive solutions to the problem. The key idea is finding the optimal sensor pose for a better online 3D reconstruction result. However, both methods often encounter difficulties in real-time planning for large areas.  When dealing with large areas, sampling-based methods \cite{bircher2016receding,zhou2021fuel,zhou2023racer,bartolomei2023multi} get more attention for they allow computation of path in real-time. However, they require well-planned spatial goals for efficient operations. The existing exploration methods focus on setting empty spaces as goals and ignore inspection-related task features. None of the existing work in goal sampling works for the swarm CARIC mission.

Coverage path planning is the process of exhaustively searching a target area with a structure of interest \cite{almadhoun2019survey}, which can be a solution to pure exploration. Prior works, such as HEDAC  \cite{ivic2023multi}, require the model of the target area for full coverage planning. Most aerial inspection service providers lack access to building models due to privacy concerns and encounter difficulties in localizing prior models with respect to local frames of reference. Consequently, model-free methods  \cite{muddu2015frontier,vemprala2018vision}  draw more attention than model-based ones.  

However, most model-free methods rely on randomized search or need to solve the Nondeterministic Traveling Salesman Problem (N-TSP), leading to low efficiency or infeasible solutions within a limited time.

The CARIC mission\footnote{\url{https://cdc2023.ieeecss.org/cooperative-aerial-robots-inspection-challenge/}} is inspired by real-life scenarios that highlight the aforementioned constraints. Existing solutions are unable to effectively handle these combined tasks and constraints, resulting in a significant domain gap that needs to be addressed. In this work, we aim to propose a robust and efficient inspection solution based on heterogeneous drones, achieving a high level of autonomy with minimal cost and time.

Our main contributions are summarized below:
\begin{itemize}
    \item We propose a task assignment method for heterogeneous UAV swarms,
    incorporating high-end LiDAR mapping drones and lower-end drones for image capturing. To the best of our knowledge, this is the first endeavor to tackle the task assignment problem for a heterogeneous UAV swarms in a realistic manner suitable for real world inspection.

    \item We introduce an efficient voxel-based multi-session coverage and balanced load navigation strategy tailored for the multi-drone inspection problem, specifically designed to address realistic limited communication conditions.

    \item We benchmark our system against state-of-the-art solutions and demonstrate superior inspection performance compared to previous methods. Additionally, we open-source our solution for the benefit of the community.
\end{itemize}

The remaining of the paper is organized as follows: Sec.  \ref{sec: Problem Description} gives a brief description of the CARIC mission. Sec. \ref{sec: overview of solution} provides an overview of our solution, whose sub-processes are described in more details in Sec. \ref{sec: implementation details}.
Sec.  \ref{sec: Experiment} presents the experiments of the proposed method and compares its performance with other existing approaches. Sec.  \ref{sec: Conclusion} concludes the whole paper.

\section{Problem Description}\label{sec: Problem Description}

The inspection mission is defined in the CARIC benchmark \cite{CARIC} based on real world scenarios. It involves a fleet of autonomous UAVs that can cooperative inspect some areas specified within bounding boxes. The UAVs can transmit photos back to a Ground Control Station (GCS) via some communication channels.
The communication among UAVs and GCS follows line-of-sight (LOS) principles to simulate real-world scenarios. A process will simulate the detection of facade defects (FDs) during the inspection missions, and the FDs are transmitted to GCS for recognition and verification. Performance evaluation is based on a total score of the FD quality, which is determined by motion blur and view angle of FD \cite{CARIC}.

CARIC provides two types of UAVs: the explorer drone (EXD) and the photographer drone (PGD). Only the EXD is equipped with a rotating 3D LiDAR to fully perceive the external environments, while both the EXD and PGD carry gimbal-stabilized cameras for inspection. 

\section{Overview of our solution} \label{sec: overview of solution}

In this section, we will provide an overview of the proposed inspection strategy for CARIC.
To make it easy to understand, the workflow is introduced through an example including two EXDs and three PGDs.

\begin{figure}
    \centering
    \vspace{-30pt}
    \includegraphics[trim={0 0cm 0cm 0cm}, clip, width=1.0\linewidth]{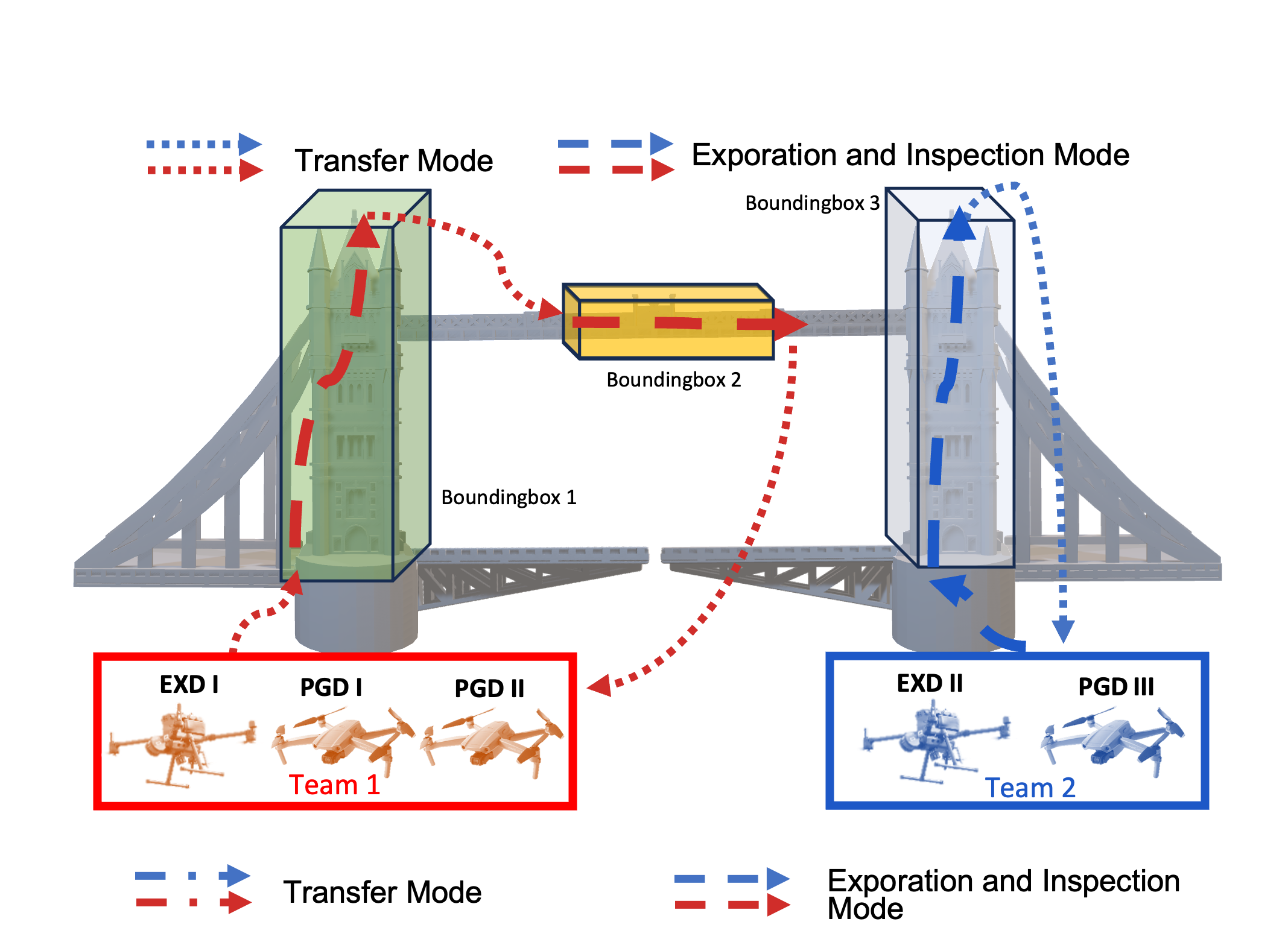}
    \caption{Illustration of how GCS assigns the \textit{task areas}: Given three bounding boxes, the GCS will balance the task areas assigned to each team based on the sizes of the bounding boxes and the teams. The UAV's activities in their task area can be under either Transfer Mode or Exploration-Inspection mode based on the primary objective.
    The details of these modes in Sec. \ref{subsec: Exploration and Inspection}.}
    \label{fig: example_begin}    
\end{figure}

The GCS will first assign the UAVs into teams based on their proximity. This so-called \textit{team assignment task} is detailed in Sec. \ref{subsec: Team and Task Assignment}. 
Then, the GCS will group some bounding boxes into \textit{task areas} and assign them to each team, as shown in Fig. \ref{fig: example_begin}.

\begin{figure*}
    \centering
    \includegraphics[width=\linewidth]{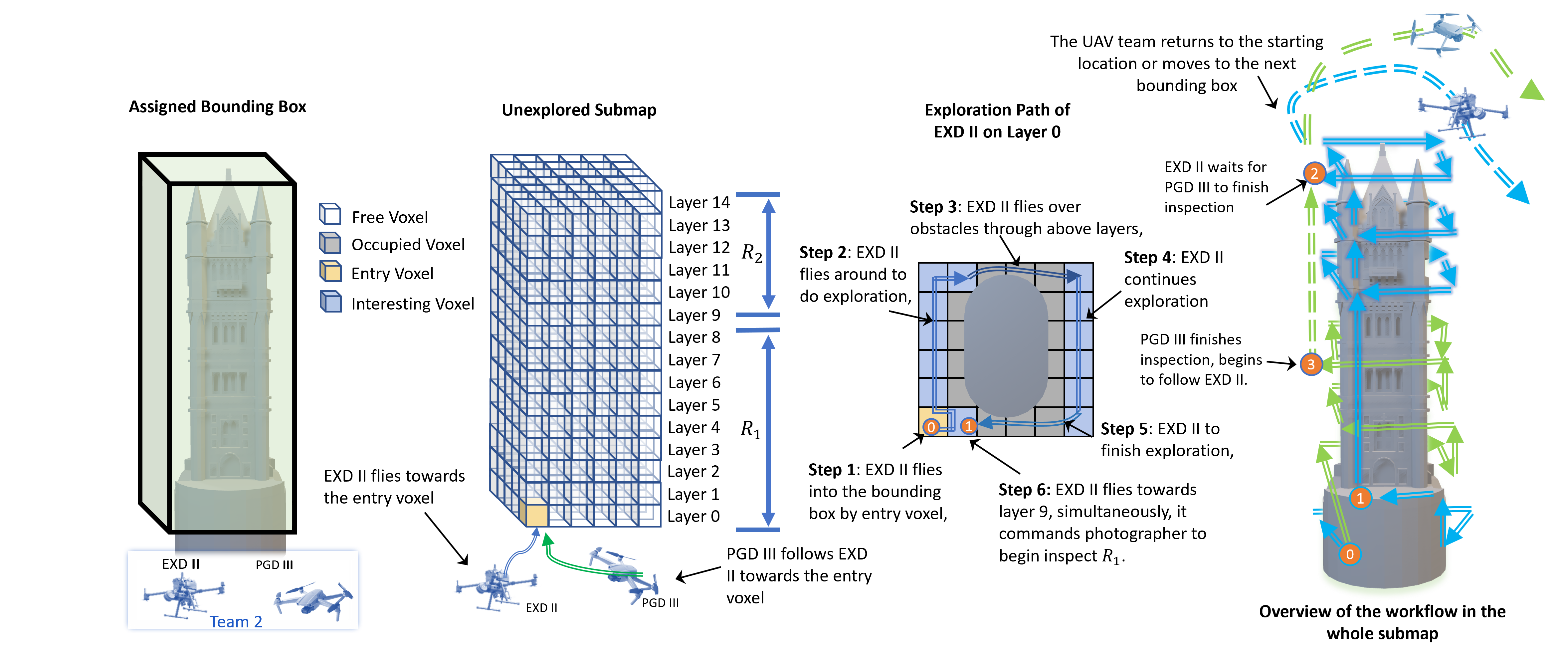}
    \caption{Example of how a team enters the bounding box and begins exploring and inspecting inside it.
    }
    \label{fig: example enter}    
\end{figure*}

Let us focus on the workflow of Team 2.
Given the task area, the UAVs will inspect the bounding boxes one by one. Each bounding box is divided into smaller layers and voxels (second part of Fig. \ref{fig: example enter}). EXD II will search for a reachable voxel on the edge of layer 1 to enter the bounding box. 
During this time, PGD III will follow EXD II's path until it enters the bounding box, as illustrated in Fig. \ref{fig: example enter}.

It should be noted that it may not be straightforward to select the entry voxel because the voxel's occupancy status is not yet known before the drone flies close enough, or other drones are occupying it; thus the EXD may have to re-select the entry voxel. The solutions to tackle these issues are discussed in Sec. \ref{sec: transfer mode}.

After entering the bounding box, the EXD first \textit{explores} the \textit{task region} $R_0$ by moving round layer 0 (the third part of Fig. \ref{fig: example enter}) and then supplies the map to a PGD (it only needs to explore the whole $R_0$ from layer 0 since the lidar can rotate around to observe the layers above). After receiving the map, the PGD will begin \textit{inspecting} the layers of $R_0$, and the EXD will simultaneously take charge of \textit{inspecting} $R_2$ (the fourth part in Fig. \ref{fig: example enter}). When a UAV finishes inspecting its asigned task region, it will wait for others to finish theirs. The \textit{exploration} and \textit{inspection} procedures will be elaborated in Sec. \ref{subsec: Exploration and Inspection}.

After all UAVs have finish inspecting the task regions, the team will move to the next bounding box in the task area. When all bounding boxes in the task area have been inspected, the UAVs will go back to their starting location, as illustrated in the last part of Fig. \ref{fig: example enter}. Note that the return to starting locations is required because the CARIC benchmark only counts the score of those FDs that have been sent back to the GCS. If a UAV inspects an area that is outside the line of sight of the GCS and does not return home, its score may not be counted at all.

\section{Implementation Details}\label{sec: implementation details}

\subsection{Map Structure}

A custom map structure is developed and serves as the foundation for the subsequent control and decision processes.
Within each team, a global map and several submaps will be maintained, where each submap corresponds to a bounding boxes and the global map is used for safe navigation from one bounding box to another.

\begin{figure}
    \centering
    \includegraphics[trim={0 0 0 0}, clip, width=1.0\linewidth]{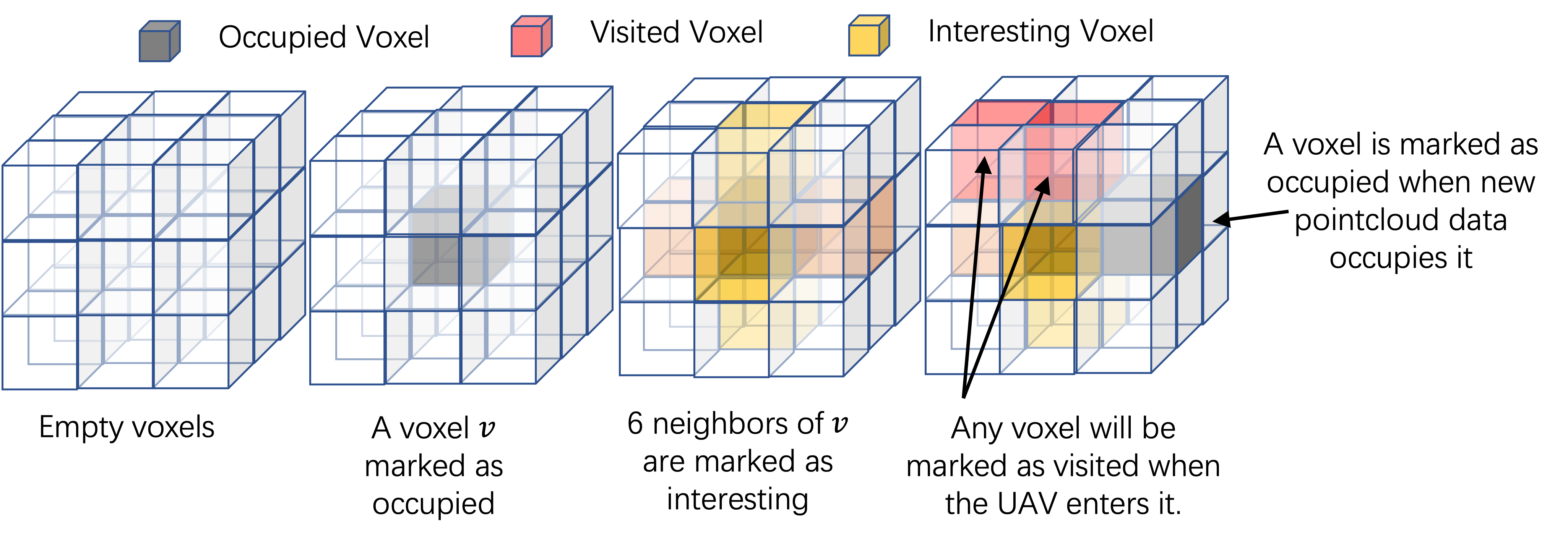}
    \caption{The voxel labelling scheme. Each voxel will have 3 boolean attributes, namely \textit{occupied}, \textit{visited}, and \textit{interesting}.}
    \label{fig: insert}    
\end{figure}

The submap consists of multiple layers, and the order of these layers follow the direction of the exit and entry points in the path $P_{T_i}$. Fig. \ref{fig: example enter} illustrates the structure of one such submap.
In the submap, each voxel has three independent boolean attributes, namely: occupied, interesting, and visited.
All voxels are initialized all attributes set as false. When a point from the input point cloud is added to a voxel, the occupied attribute is set to True, and the six side-adjacent neighbours of $V$ will have the interesting attributed set to True. Then the UAV will sequentially move to each \textit{unoccupied and interesting} voxel in the layer during the exploration phase.
When a UAV enters any voxel, the visited attributed will also be set to True.
The process is elaborated in detail in Fig. \ref{fig: insert}.

\subsection{Team Assignment}\label{subsec: Team and Task Assignment}

To avoid solving a complicated N-TSP, and since the number of UAVs is small, a simple rule-based approach is designed to group EXD and PGD into teams.
Spcifically, the GCS initially identifies the number of EXDs present within the environment (assuming that all drones are initially in LOS to the GCS). 
Teams are formed, with each team consisting of a single EXD as the leader. 
Then, PGDs are assigned to teams based on the Euclidean distance to the respective EXD.

\subsection{Task Area Assignment} \label{sec: task area}

Following the team assignment task, the GCS will find a path going through all $N$ the bounding boxes, Specifically, the for bounding box $B_i$, $i \{1, \dots N\}$, we find the principle axis $a_i$ that goes through the center and aligns with the longest dimension of $B_i$. Then we find the pair of points $(p_{1}^i$, $p_{2}^i)$ which are the intersections of $a_i$ with the two sides of the bounding box (indeed, $p_{1}^i$, $p_{2}^i$ are also the center of these sides). These so-called enter-exit pairs are grouped together into a single set $S = \{(p_{1}^i, p_{2}^i), i = 1, \dots N\}$.

Next, we try to find the shortest path $P = (p_1, p_2, \dots p_{2j-1}, p_{2j}\dots p_{2N})$ such that for each $j \in \{1, N\}$, $(p_{2j-1}, p_{2j}) \in S$ or $(p_{2j-1}, p_{2j}) \in S$. Moreover $p_1$ is the closest point to the GCS. To find an approximate of the shortest path, we adopt a simple first-best approach.
When the path $P$ has been determined, we will assign a sub-path $P_{T_i}$ of $P$ to team $T_i$ based on the size of the team and the volume of the bounding boxes that $P_{T_i}$ goes through. We refer to these bounding boxes as the \textit{task area}.

\begin{figure}
    \centering
    \includegraphics[trim={0 0 0 0}, clip, width=1.0\linewidth]{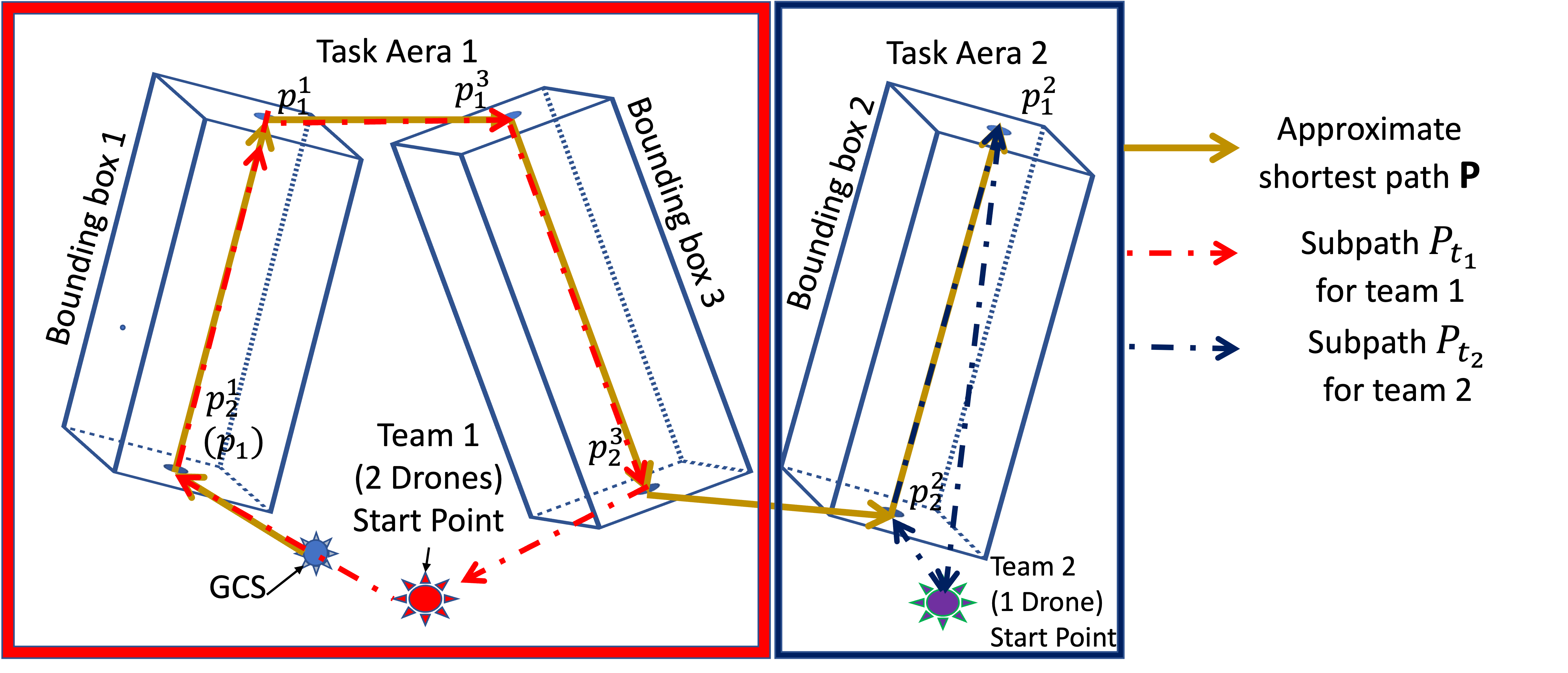}
    \caption{An example of task assignment: a pair of point $(p^i_1, p^i_2)$ is identified for each bounding box, then an approximate of the shortest path $P$ is calculated by the first-best approach. Then subpaths of $P$ will be assigned to the teams based on the bounding box volumn.}
    \label{fig: task assign math}    
\end{figure}

\subsection{Exploration and Inspection}\label{subsec: Exploration and Inspection}

After initializing the map, each team will begin the exploration and inspection process. This process has two modes, namely, Transfer Mode and Exploration-Inspection Mode.

\subsubsection{Transfer Mode} \label{sec: transfer mode}

\begin{figure}
    \centering
    \includegraphics[width=\linewidth]{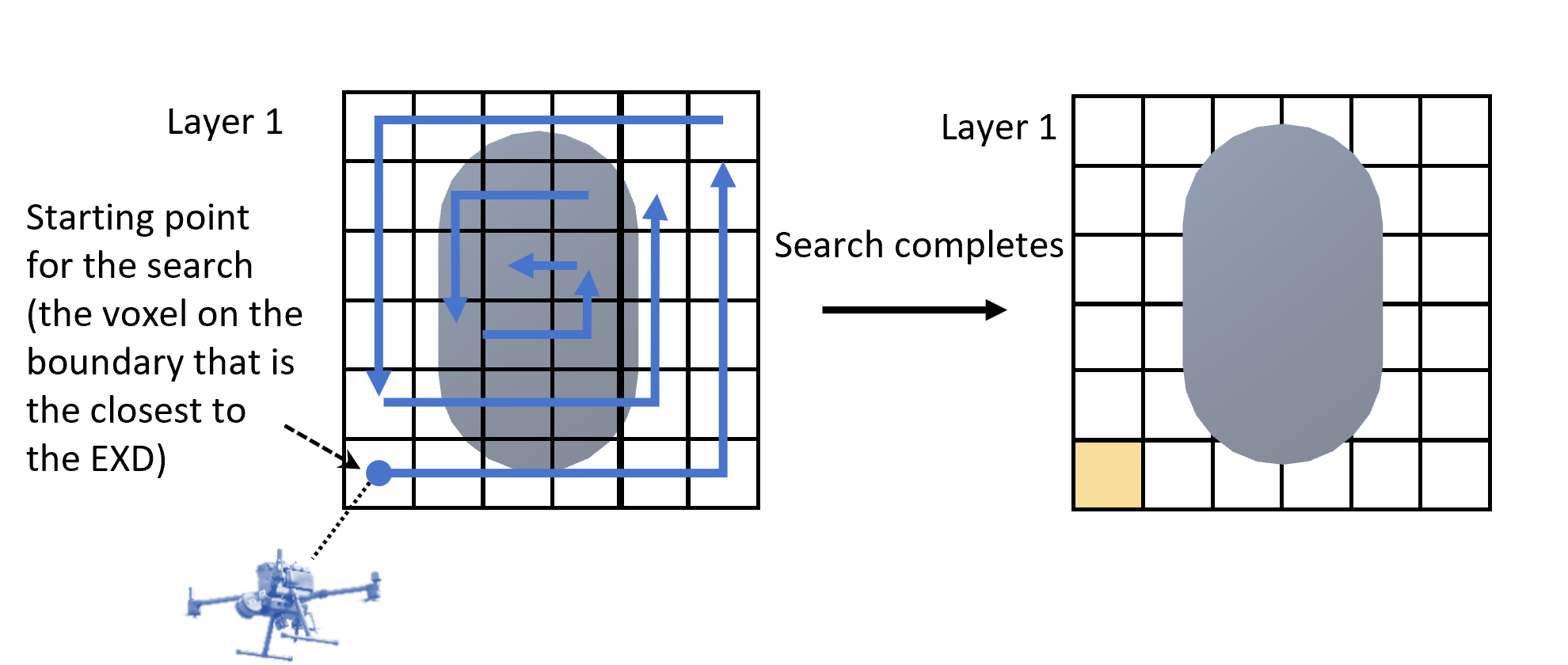}
    \caption{The EXD will search for an entry voxel to the submap in a spiralling pattern, starting from the closest voxel.}
    \label{fig: enstry search}
\end{figure}

This mode is carried out in between the exit of a submap and the entry of another in the task area.
First,
the EXD has to determine and reach an entry voxel in layer 0 of the target submap. Fig. \ref{fig: enstry search} shows the sequence that the voxels will be evaluated to become the entry point, starting from the voxel that is closest to the EXD. For each voxel on this sequence, we will invoke the A$^*$ algorithm to find a path to it, comes in the form of a sequence of position setpoints.
During the this process, occupancy of the voxels in both global and submaps will be continuously updated as new point clouds are received. If the UAV can reach one of voxels, it will automatically become the entry point and the system transitions to the Exploration and Inspection Mode. If A$^*$ determines that the voxel is not longer reachable, we change to evaluating the next voxel in the sequence. If none of the voxel in the first layer can become entry point, then we move on the evaluate the next layer.

During the Transfer Mode, the PGD continuously tracks the EXD movements. Once the EXD exits the Transfer Mode, 
the PGD navigates to the designated entry voxel and wait for further instructions from EXD. 
This coordinated interaction ensures that the PGD aligns seamlessly with the EXD actions, maintaining efficient and secure navigation throughout the transition from Transfer Mode to subsequent Exploration and Inspection Mode.

\subsubsection{Exploration and Inspection Mode}

\begin{figure}[ht]
    \centering
    \includegraphics[trim={0 0 0 10}, clip, width=1.0\linewidth]{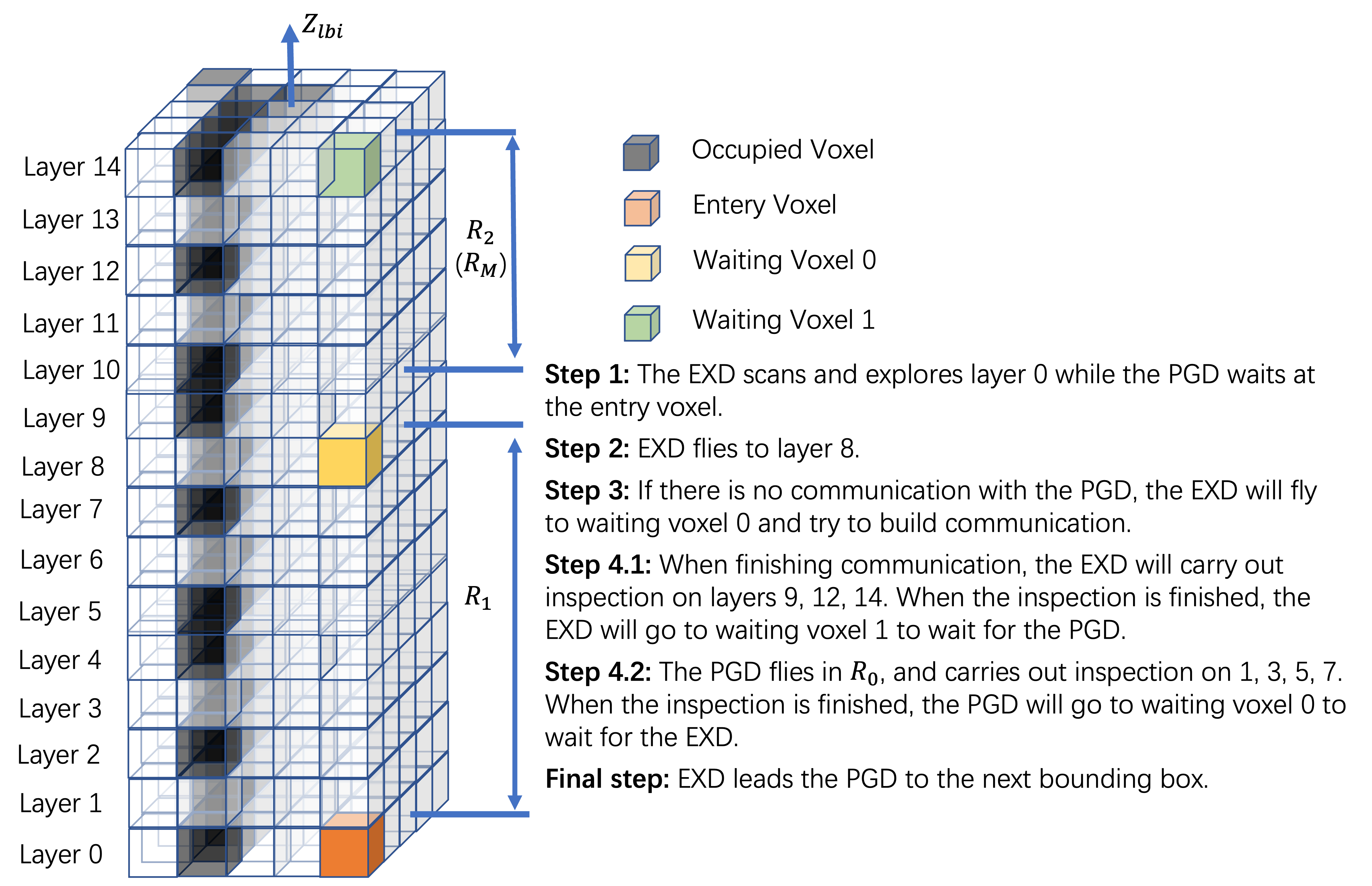}
    \caption{Details of a team with one EXD and one PGD exploring and inspecting one assigned bounding box.}
    \label{fig: example_rule}    
\end{figure}

Once the EXD enters the entry voxel, the Exploration and Inspection Mode will commence.
At the beginning of this mode, the layers in the submap are divided into multiple regions $R_1, R_2, \dots R_M$, where $M$ is the number of UAVs in the team.
The EXD will explore the regions $R_1,\dots R_{M-1}$ before handing them over to the PGDs for further inspection. In each region EXD undertakes a sequence of actions:
\begin{itemize}
    \item  Quickly mapping the first layer in the region:
    To obtain an occupancy map of a region, the strategy used is to fly the EXD around the bottom layer (since the lidar of CARIC can rotate periodically).
    The objective here is to visit the \textit{interesting}, \textit{unvisited} voxels within the bottom layer. To do so, we use a modified Dijkstra algorithm. If the bottom layer does not have occupied voxels, i.e. no interesting voxel, the search returns the voxel at the boundary that remains unvisited in this layer. 
    \item  Supply the map to the PGD: Upon successful completion of the mapping, the EXD will proceed to the bottom layer of the next region. Simultaneously, it will send a command to a PGD to begin inspecting the region by the EXD. The EXD will actively maintain communication with the PGD while it's working on the assigned region. 
    \item  Communication handling: In instances where the command to PGD cannot be issued due to lack of LOS communication, the EXD will try to navigate to a location in LOS of PGD, (step 3 in Fig. \ref{fig: example_rule}),  which waits at the entry voxel. If this approach is unsuccessful, the EXD continues by moving to the adjacent free voxels.

\end{itemize}
This series of actions enables the EXD to synchronize with the PGD and accommodate communication challenges that may arise during the mission.\\

After mapping the regions $R_1\dots R_{M-1}$ and assigning the PGDs to explore them, the EXD proceeds to conduct a detailed inspection of $R_M$.
Based on the provided camera field of view (FOV) and scoring metrics, EXD only needs to explore one out of every three layers.
To explore a target layer, the EXD uses the Dijkstra search to navigate to unvisited but interesting voxel at that specific layer. If the search no longer finds a target, the EXD progresses to the next target layer. Upon completing all assigned tasks, the EXD navigates to the a waiting voxel (see Fig. \ref{fig: example_rule}) and assumes the waiting state.

For the PGDs, upon receiving the command indicating that the EXD has mapped the tasked subregion, the PGD undertakes a series of actions as following steps:
\begin{itemize}

    \item The PGD, prompted by the command, enters the designated task area via the entry voxel.

    \item Inspection: Similar to the EXD, within the task area, the PGD organizes the inspection process on one of every three layers in its assigned region. The PGD employs the Dijkstra algorithm to visit each layer thoroughly.
    
    \item Transition to waiting State: Upon completing the inspection of all designated layers, the PGD proceeds to navigate to the nearest voxel that shares the same x-index and y-index as the entry voxel. In this location, the PGD assumes a waiting state and awaits further instructions from the EXD.
    
\end{itemize}

When all the PGDs have finished their assignments, the EXD commands all team members to transition into Transfer Mode. The team collectively takes flight towards the next bounding box area. 

\subsection{Collision Avoidance Rules}\label{subsec: Collision Avoidance Rules}
To ensure effective collision avoidance, all agents within the system publish their current global positions and future positions according to the planned paths. This information is shared among the agents and serves as a reference during their path-planning processes. Different obstacle avoidance priorities are assigned to agents based on their roles and the alphabetical order of their names to prevent deadlock issues arising from mutual obstruction between agents. This collaborative approach enhances the overall safety and efficiency of the system's operations.

\section{Experiment}\label{sec: Experiment}
We conducted various experiments to test our proposed solution. Due to the page limit, we only show results for the following aspects:
\begin{enumerate}
\item Single-UAV inspection scenario;
\item Multi-UAV inspection scenario;
\item Ablation study for the efficacy of the swarm.
\end{enumerate}
The experiment scenario is shown in Fig.\ref{fig: mission_describe}, consists of an iconic Marina Bay Sand model and a bounding box of size 130m $\times$ 70m $\times$ 60m, which is larger than most of existing full coverage research with no prior. For simplicity, we use the fleet structure to describe the proposed method, for example, 1E0P means there is one EXD and no PGD here to do this mission. 

The methods are evaluated based on the score obtained during the mission progress, which is computed according to the metrics introduced in \cite{CARIC}.
\subsection{Single-UAV Experiment}\label{Single-UAV Experiment}
\begin{figure}[ht]
    \centering
    \includegraphics[trim={0 0cm 0cm 0cm}, clip, width=0.96\linewidth]{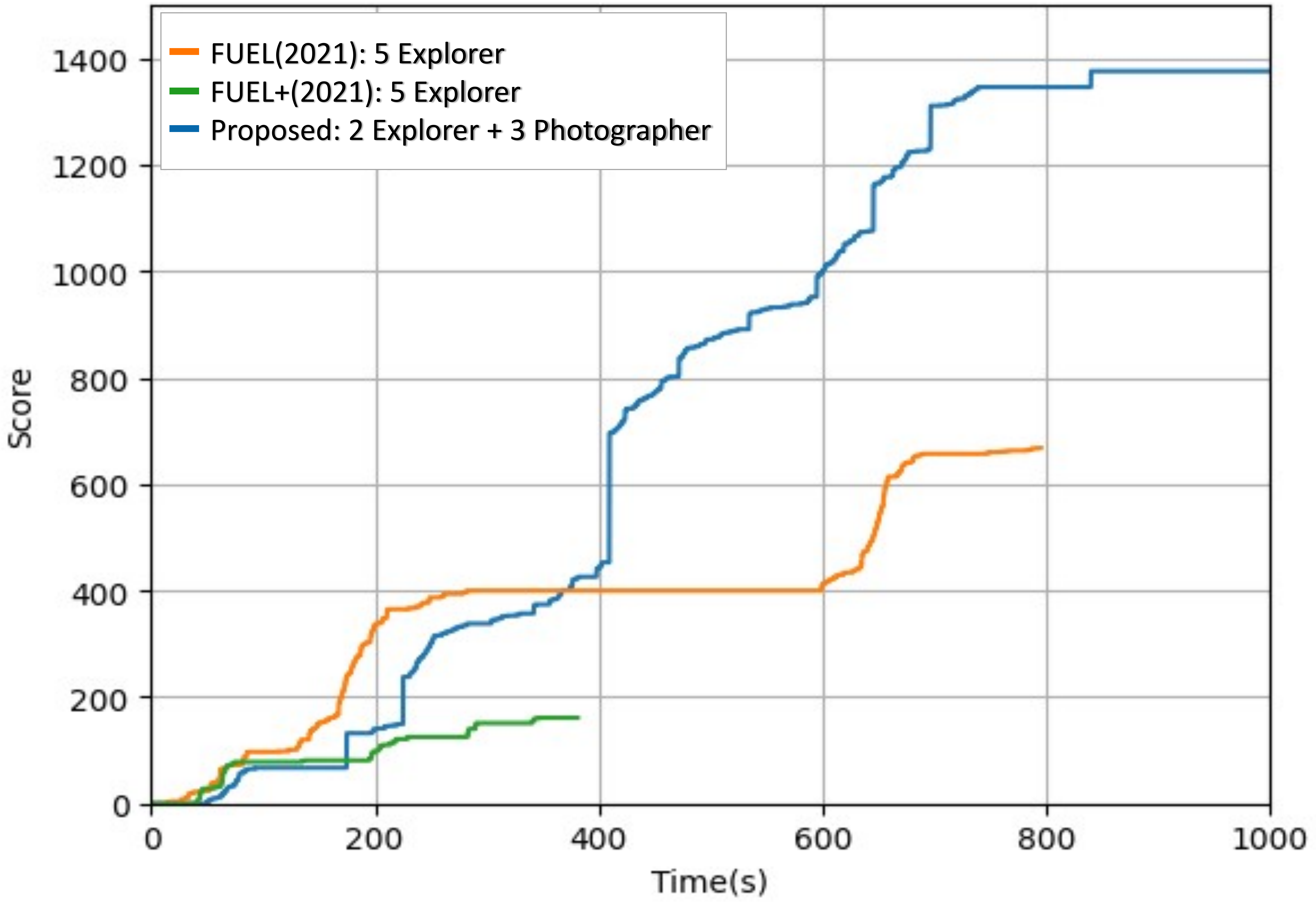}
    \caption{The inspection score of the single-UAV experiment }
    \label{fig: single-uav  result}    
\end{figure}

In the single-UAV experiment, we compare our method with the single-UAV exploration method FUEL \cite{zhou2021fuel}. 
The FUEL algorithm is not directly applicable for large outdoor missions due to its limitations: (1) it assumes no communication constraints, and (2) it is designed for use with a 5-meter range, 90-degree field of view (FOV) RGBD camera inputs for a small indoor environment. Directly using FUEL could result in mission failure due to a lack of communication or the loss of track of position or obstacles in large scenes, as shown in Fig. \ref{fig: single-uav  result}.\\
\begin{figure}[ht]
    \centering
    \includegraphics[trim={0 0cm 0cm 0cm}, clip, width=0.96\linewidth]{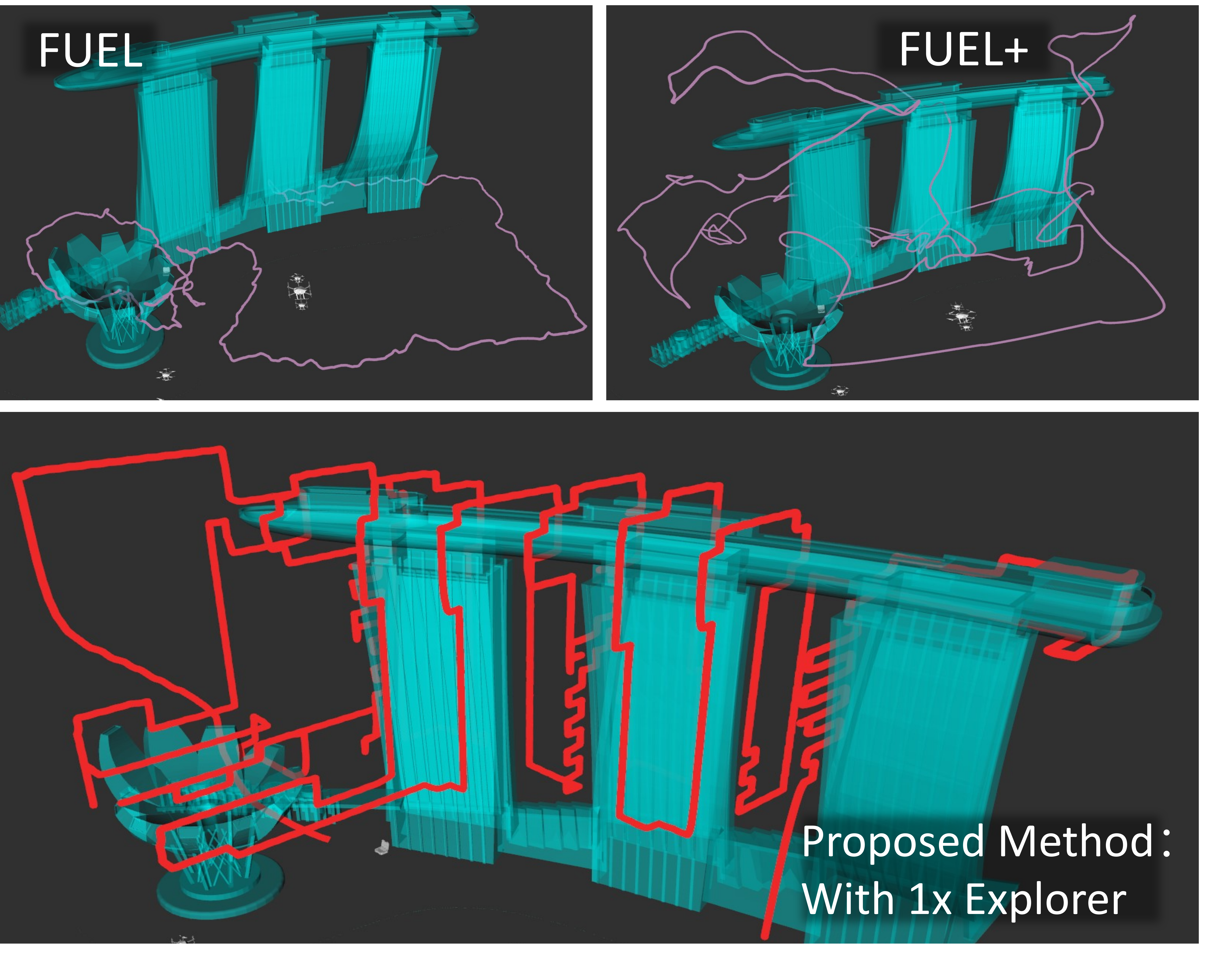}
    \caption{The trajectory in single-UAV experiment }
    \label{fig: single-uav traj}    
\end{figure}
To overcome the limitations of the FUEL method, we simplify the problem for FUEL by giving FUEL 100\% communication coverage even without LOS, while our proposed solution follows LOS communication constraints. Additionally, we extend the observation range of the 90-degree dense RGBD model to the same as that of the Lidar sensor, allowing for observation of hundreds of meters in RGBD image format. While this model may not be realistic, it allows for a more fair comparison. For the modified observations, we refer to this FUEL variant as FUEL+ due to the increased depth range, as shown in Fig. \ref{fig: single-uav  result}.\\
However, despite these advantages we give to the algorithm, FUEL+ is still unable to cover the entire environment due to its inherent limitations in observation FOV and lack of robustness for searching multiple height layers. Both FUEL and FUEL+ prematurely terminate the mission, believing the mission is completed and there is no more place to search, as shown in Fig. \ref{fig: single-uav traj}. As a result, our solution beats FUEL and FUEL+ by a large margin. Detailed analysis revealed that FUEL and FUEL+ are more suited for single 2D layer exploration, similar to 2D LiDAR-based exploration methods in previous research. This renders them less effective for modeling high-rise buildings and buildings with complex architectural styles. While most people anticipate that denser image formats would enhance exploration and be ideal for integration with AI, this experiment suggests otherwise, indicating that wider but sparsely covered sensors are more critical for making accurate exploration and coverage decisions.

\subsection{Multi-UAV Experiment}\label{Multi-UAV Experiment}
\begin{figure}[ht]
    \centering
    \includegraphics[trim={0 0cm 0cm 0cm}, clip, width=1\linewidth]{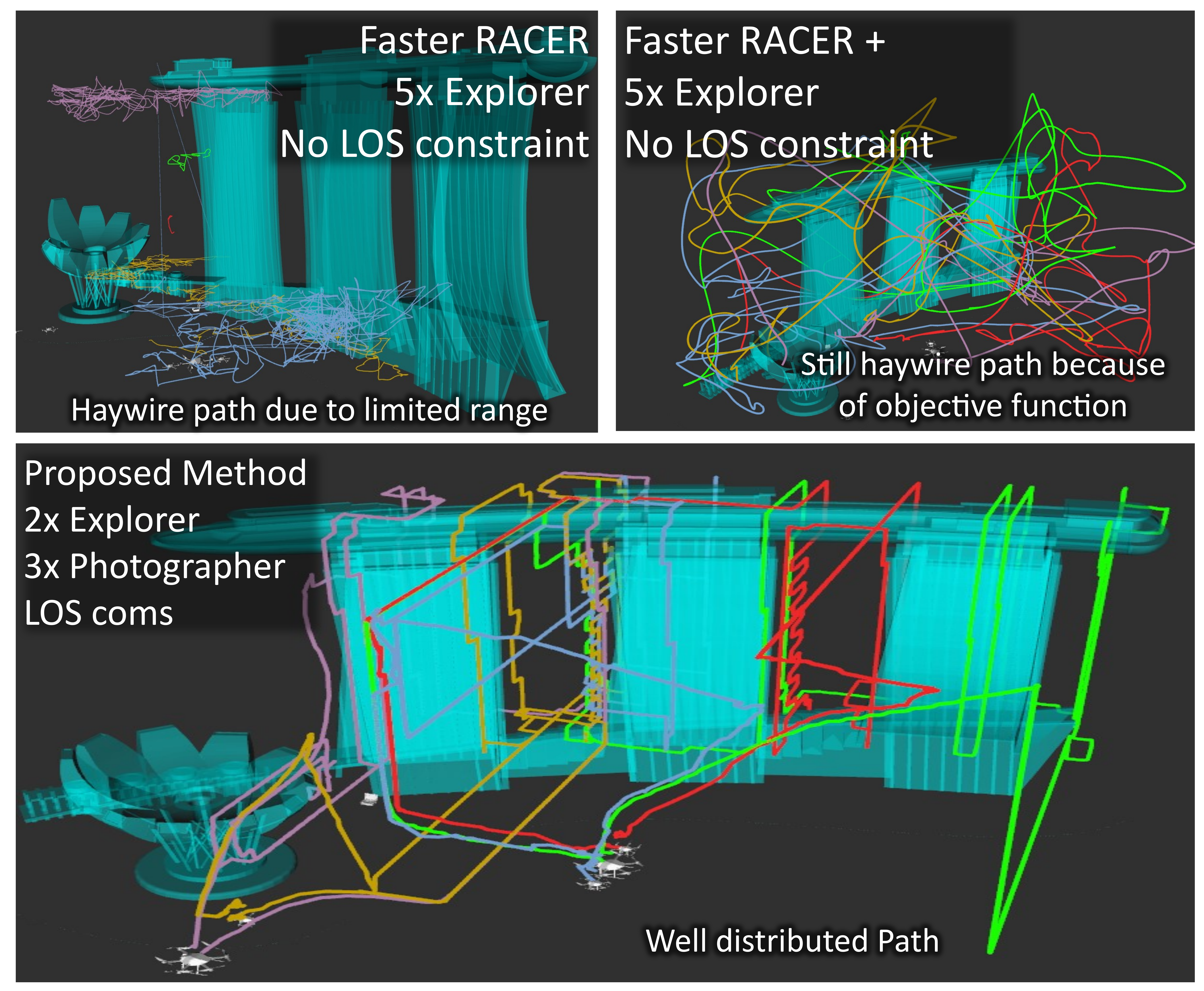}
    \caption{The trajectory in Multi-UAV experiment }
    \label{fig: multi-uav traj}    
\end{figure}

\begin{figure}[ht]
    \centering
    \includegraphics[trim={0 0cm 0cm 0cm}, clip, width=1\linewidth]{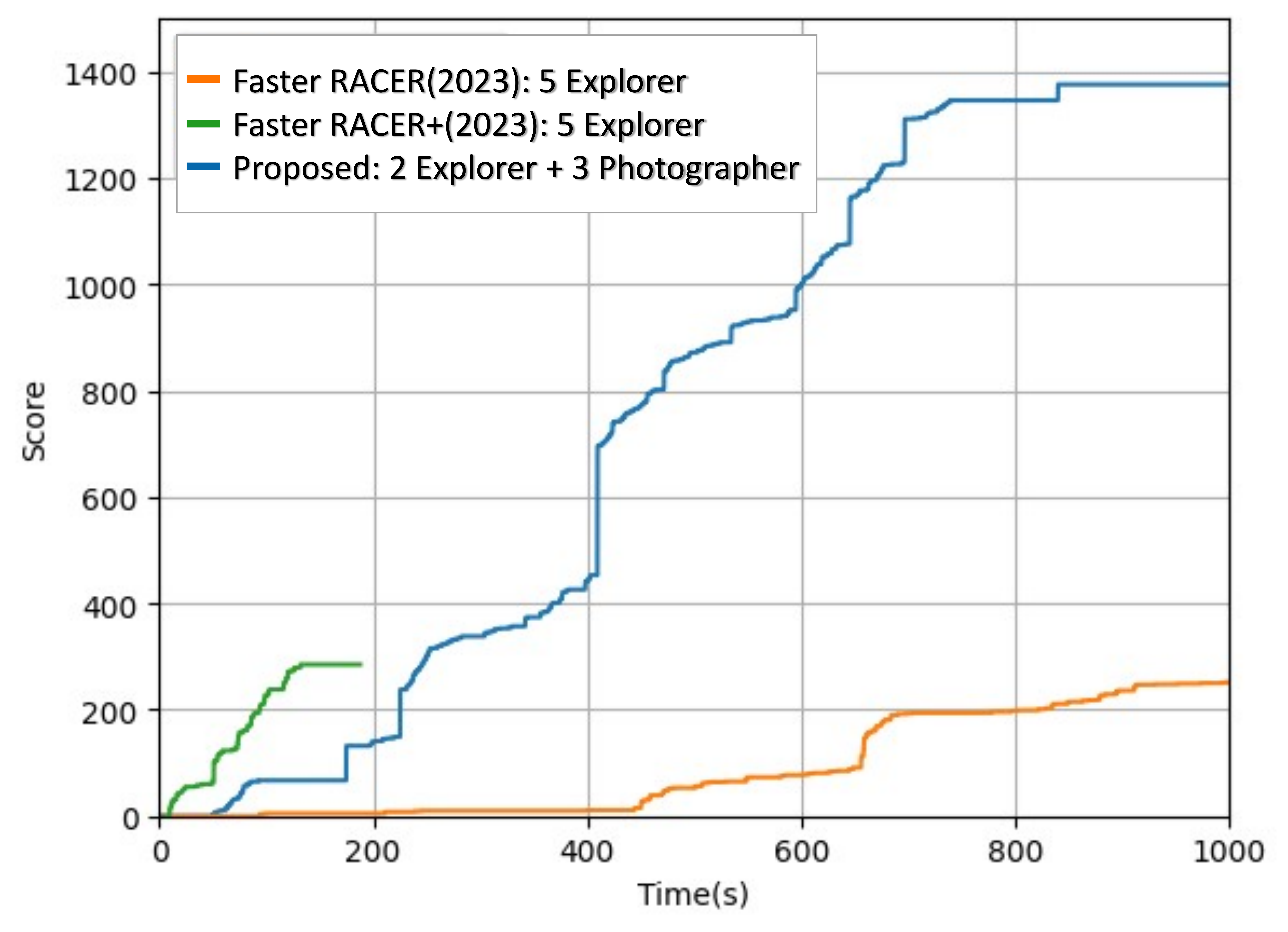}
    \caption{The results of Multi-UAV experiment }
    \label{fig: multi-uav results}    
\end{figure}

In the multi-UAV experiment, we compare our method with SOTA Faster RACER  \cite{bartolomei2023multi}, the newest benchmark in autonomous exploration. Similar to the approach with FUEL, we also introduce Faster RACER+, an enhanced version of Faster RACER with ideal sensors and unlimited communication, to let RACER start running. The result is shown in the Fig.\ref{fig: multi-uav results}. Referring to the trajectory of the UAV shown in Fig.\ref{fig: multi-uav traj}, Faster RACER+ indeed demonstrates exploration efficiency in the early stage of the mission. However, due to the gap between the optimal inspection range and the optimal exploration range, Faster RACER+ fails to focus on detailed structural information, which typically requires closer sensor proximity for better observation.
Compared with FUEL, although Faster RACER+ has decentralized workload assignments for exploring the entire environment, the issue of slow exploration for areas of interest in large environments still persists. The proposed method can complete inspection tasks under challenging conditions, such as limited communication, at least twice as fast as the benchmark.

\subsection{Ablation study for the efficacy of the swarm}\label{Robustness Experiment}
\begin{figure}[ht]
    \centering
    \includegraphics[trim={0 0cm 0cm 0cm}, clip, width=1\linewidth]{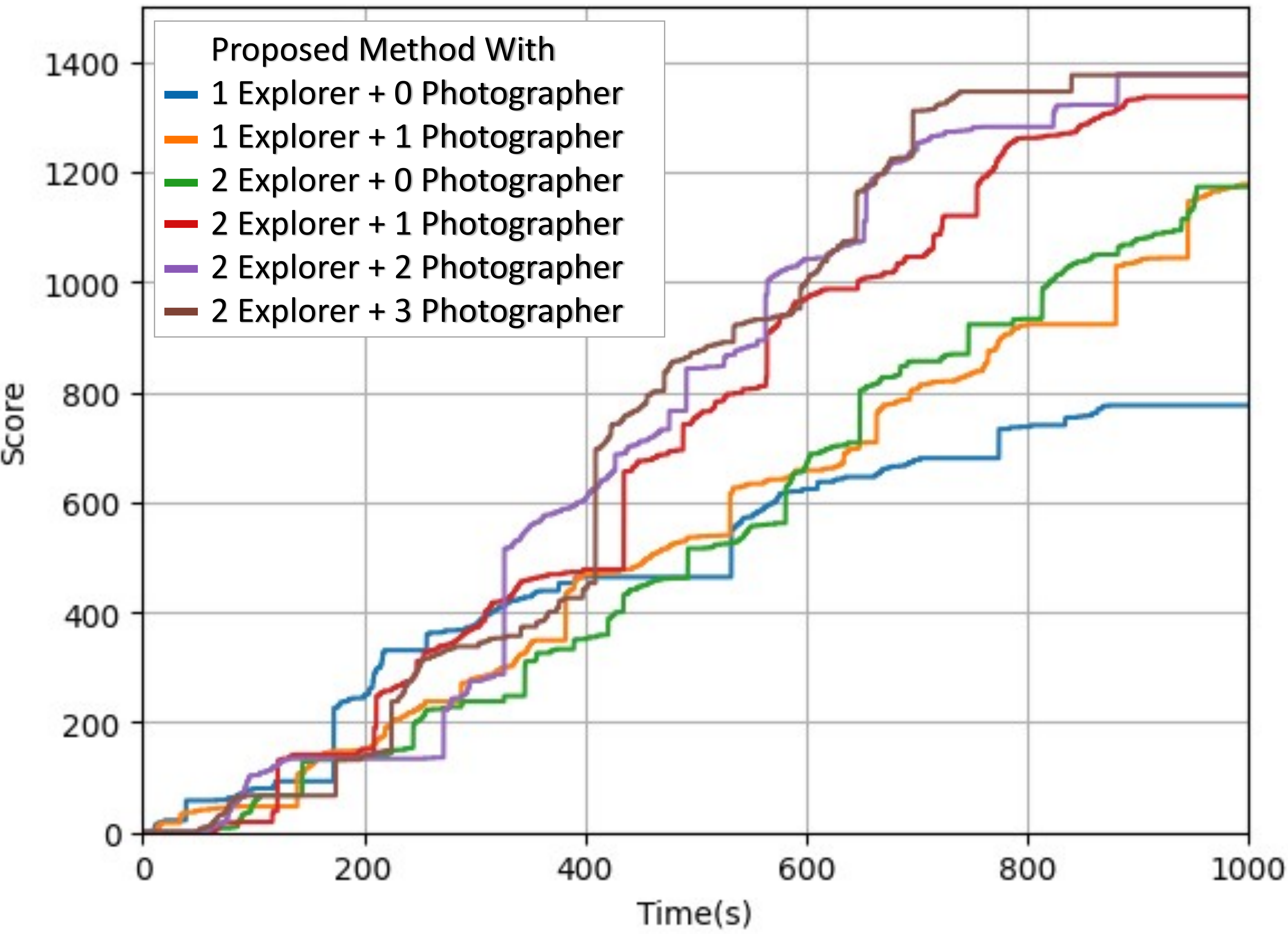}
    \caption{Ablation study for the efficacy of the swarm.}
    \label{fig: robustness result}    
\end{figure}

To evaluate the efficacy of the proposed swarm solution, we conduct tests on identical inspection scenarios with varying swarm sizes and task force sizes, as shown in Fig. \ref{fig: robustness result}. In the experiment, performance improves with an increasing number of UAVs. 
Efficiency gains are noticeable when the total drone count is three or fewer but become less apparent as the fleet size exceeds three. For a test environment of 130mx70mx60m, four drones can already achieve good coverage. Increasing the number of drones also means higher perceptions of overheads. To better differentiate the performance, a larger test environment size is needed. Although the proposed method may not achieve the ideal of \textit{1+1$>$2}, the experiment demonstrates its robustness and stability. The scalability of the method is shown, along with improved coverage speed.

Moreover, we have successfully verified its environmental adaptive ability by testing the method in different scenes, such as the height-limited sense \textit{hangar} and the complex vertical space like \textit{crane} for the actual CARIC competition. 
\section{Conclusion}\label{sec: Conclusion}
In this paper, we propose a cost-effective strategy for heterogeneous multi-agent systems to do cooperative aerial inspection in partially known environments. By single-UAV experiment in section \ref{Single-UAV Experiment} and multi-UAV experiment in section  \ref{Multi-UAV Experiment}, the superiority of the proposed algorithm is demonstrated. In the robustness experiment in section \ref{Robustness Experiment}, we tested our algorithm on a variety of drone combinations to verify the robustness of our algorithm. In summary, we fill the gap of cooperative exploration and inspection strategy for heterogeneous aerial systems.  Additionally, we open-
source our solution to contribute to the research community.

\balance
\bibliographystyle{ieeetr}
\bibliography{reference}

\end{document}